\documentclass[letterpaper, 10 pt, conference]{ieeeconf}  % Comment this line out if you need a4paper

\usepackage{tikz}
\usepackage{float}
\usepackage{subcaption}
\usepackage{amsmath}
\usepackage{algorithm}
\usepackage{algpseudocode}
\usepackage{hyperref}
\usepackage[sorting=none, backend=bibtex]{biblatex}

\newcommand\copyrighttext{%
	\footnotesize \textcopyright 2020 IEEE. Personal use of this material is permitted.
	Permission from IEEE must be obtained for all other uses, in any current or future
	media, including reprinting/republishing this material for advertising or promotional
	purposes, creating new collective works, for resale or redistribution to servers or
	lists, or reuse of any copyrighted component of this work in other works.
}
\newcommand\copyrightnotice{%
	\begin{tikzpicture}[remember picture,overlay]
	\node[anchor=south,yshift=10pt] at (current page.south) {\fbox{\parbox{\dimexpr\textwidth-\fboxsep-\fboxrule\relax}{\copyrighttext}}};
	\end{tikzpicture}%
}

\bibliography{quellen}
\AtBeginBibliography{\small}

\IEEEoverridecommandlockouts                              % This command is only needed if 
\title{\LARGE \bf
OpenREALM: Real-time Mapping for Unmanned Aerial Vehicles
}

\author{Alexander Kern$^{1}$, Markus Bobbe, Yogesh Khedar and Ulf Bestmann% <-this % stops a space
}

\begin{document}

\maketitle
\copyrightnotice

\thispagestyle{empty}
\pagestyle{empty}

%%%%%%%%%%%%%%%%%%%%%%%%%%%%%%%%%%%%%%%%%%%%%%%%%%%%%%%%%%%%%%%%%%%%%%%%%%%%%%%%
\begin{abstract}

This paper presents OpenREALM, a real-time mapping framework for Unmanned Aerial Vehicles (UAVs). A camera attached to the onboard computer of a moving UAV is utilized to acquire high resolution image mosaics of a targeted area of interest. Different modes of operation allow OpenREALM to perform simple stitching assuming an approximate plane ground, or to fully recover complex 3D surface information to extract both elevation maps and geometrically corrected orthophotos. Additionally, the global position of the UAV is used to georeference the data. In all modes incremental progress of the resulting map can be viewed live by an operator on the ground. Obtained, up-to-date surface information will be a push forward to a variety of UAV applications. For the benefit of the community, source code is public at \href{https://github.com/laxnpander/OpenREALM}{https://github.com/laxnpander/OpenREALM}.

\end{abstract}

%%%%%%%%%%%%%%%%%%%%%%%%%%%%%%%%%%%%%%%%%%%%%%%%%%%%%%%%%%%%%%%%%%%%%%%%%%%%%%%%
\section{Introduction}
\label{sec_intro}
The surveying of the world has always been an elementary part of human curiosity. With invention of the aircraft and the success of commercial aviation in the past century an additional viewpoint has been added to this curiosity. Observing the world from above never fails to impress. But it is not only an astonishing perspective, aerial imagery has made its way into diverse areas of life. Precision agriculture, urban planning and disaster risk management are only some of the most promising topics in the upcoming years, that will profit from advancements in this field \cite{global_rep}. 
% Bisschen cheasy insgesamt, aber vll ja motivierend?

Besides resolution of the acquired data, several other factors will thereby define the success. Availability, costs and time of the mapping process are of crucial importance to almost any of the use cases. The rapid progress in the field of Unmanned Aerial Vehicles (UAVs) bears a chance to significantly contribute to a solution for these challenges. Thanks to their potentially lightweight design and small dimensions UAVs are quickly manufactured in large numbers. With the development of open source autopilots their autonomy has reached a level, where almost any briefly trained user can safely plan and execute flight missions. This trend already democratised aerial imagery and led to numerous software for processing and evaluation of acquired aerial data, such as Agisoft Metashape \cite{Agisoft} or DroneDeploy \cite{DroneDeploy}. 
The technique behind such software is typically referred to as 'photogrammetry'. While in the classical understanding any kind of spatial measurement considering shape or position of objects in images is referred to as photogrammetry, today a very specific workflow is understood under this term. It involves algorithms such as bundle adjustment, that compute highly accurate camera pose and 3D surface information of the observed scene in a large, geometric optimization formulation.
% Vll noch eine bsp software rauswerfen, damit absatz besser passt?
\begin{figure}[H] % Bilder vielleicht noch gerade rücken? oder ne kompassrose drauf?
	\centering
	\begin{subfigure}[b]{\linewidth}
		\centering
		\includegraphics[height=7cm,keepaspectratio]{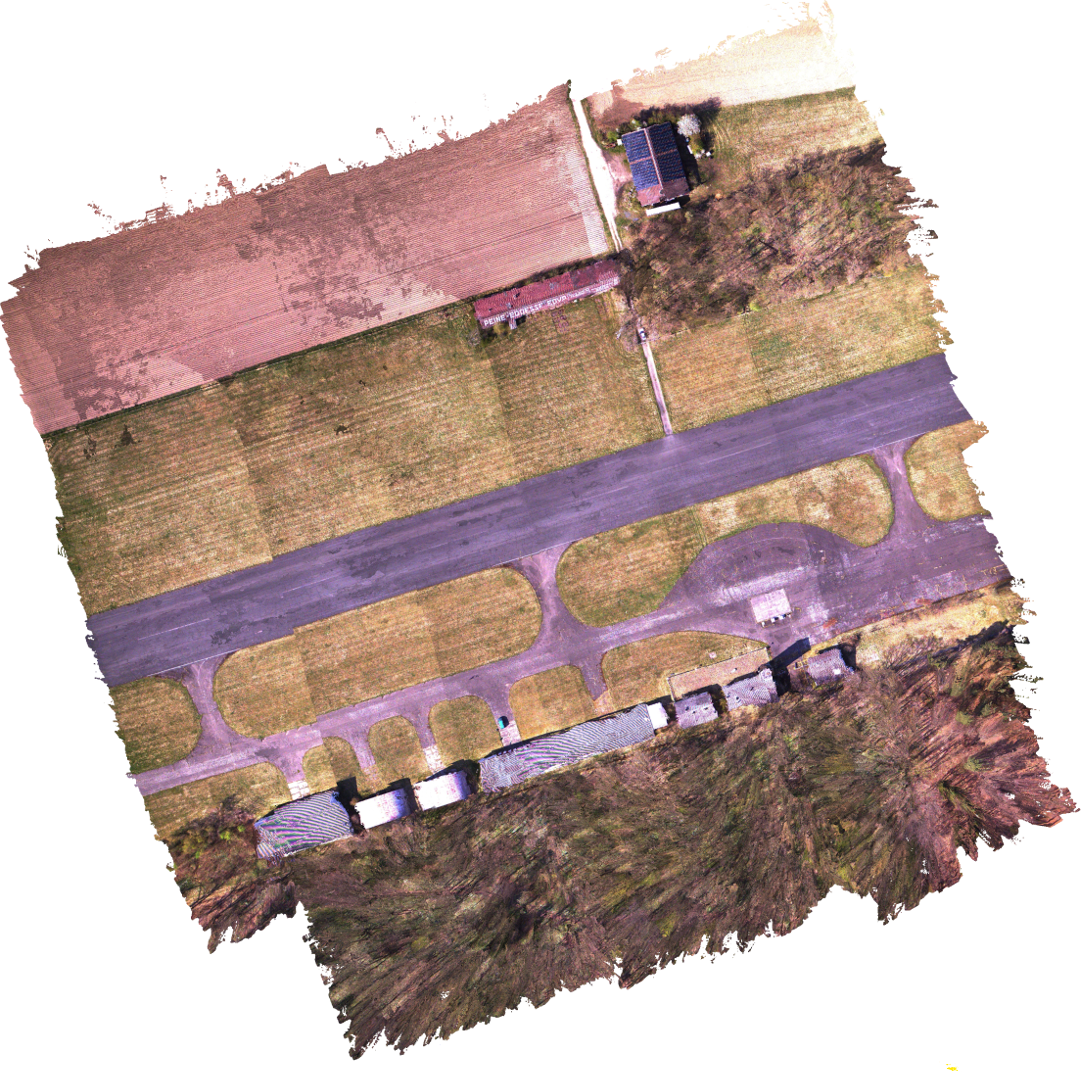}
		\caption{The computed map is georeferenced and corrected by geometric distortions induced by elevation variations, therefore the result is an incrementally updated, true orthophoto.}
	\end{subfigure}
	\begin{subfigure}[b]{\linewidth}
		\centering
		\includegraphics[height=7cm,keepaspectratio]{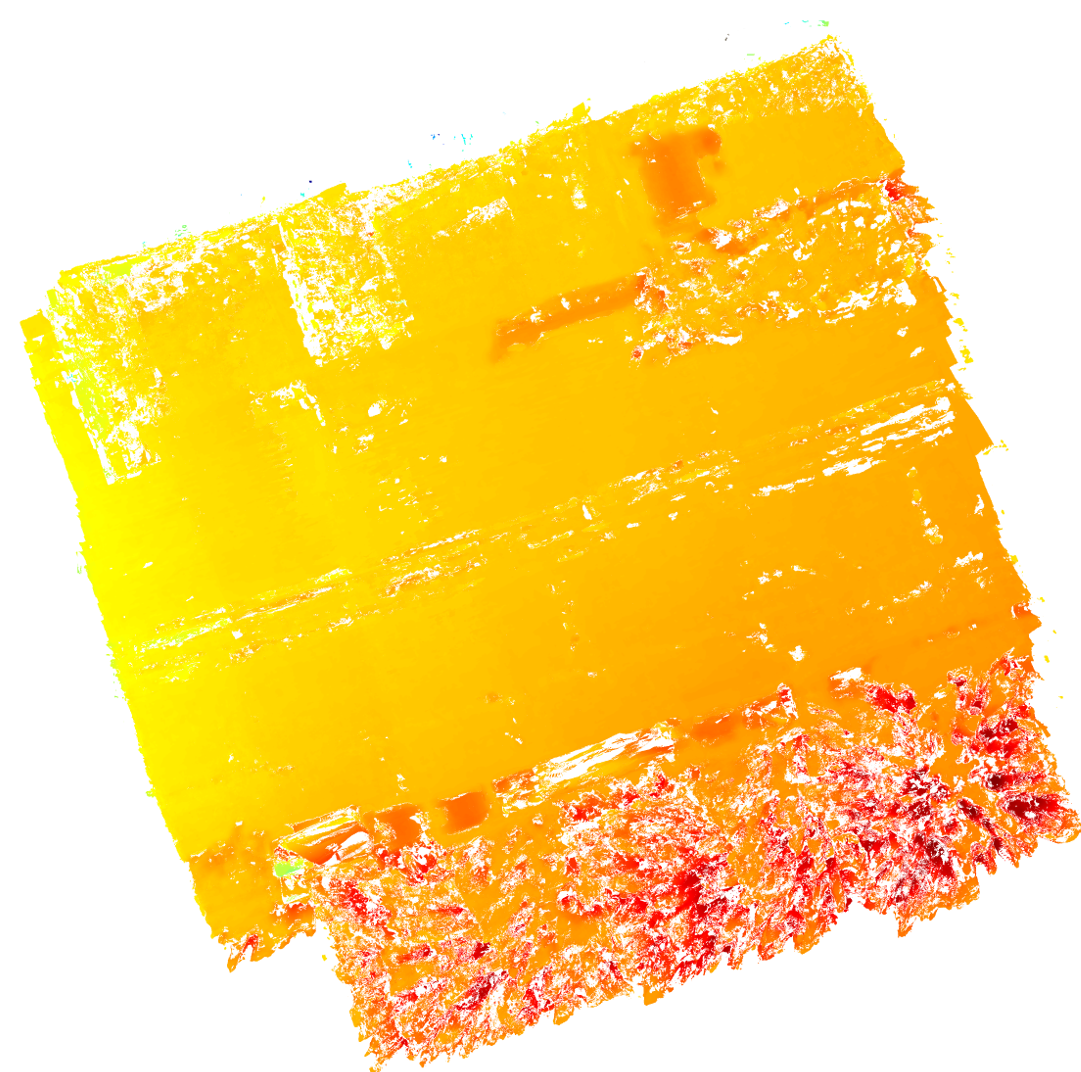}
		\hfill
		\caption{Ground surface is represented as elevation map generated from 3D point cloud data. A probabilistic approach identifies data regions with high noise and removes them from the view.}
	\end{subfigure}
	\caption{OpenREALM operates in real-time, while the UAV is still in the air. It relies on state-of-art visual pose estimation and monocular dense reconstruction frameworks to fully reconstruct surface information.}
	\label{fig:ortho_elevation}
\end{figure}
Despite bundle adjustment being exceptionally efficient and fast considering the usually large number of parameters to be refined \cite{Triggs00bundleadjustment}, most of the mentioned frameworks exclusively work after flight with all images captured. This is not optimal in several ways. Intervention during the mission based on the acquired data is not possible, although all information is theoretically available. The overall time that passes from mission start to finalization of the global map can take up to several hours, which in case of a search and rescue scenario is not acceptable. And finally the acquisition time is not used for processing, even though the UAV is operating autonomously.

Within the scope of this paper we therefore propose a framework, that is capable of reading a continuous image stream acquired by a calibrated, downward pointing, 2-axis gimbal stabilized camera attached to a UAV. It offers four different modes of operation, each increasing in complexity generating
\begin{enumerate}
	\item 2D maps from generic images (RGB, IR, ...) based on GNSS position and heading,
	\item 2D maps from RGB images based on GNSS position and visual pose estimation,
	\item or 2.5D digital elevation models from RGB images based on GNSS positioning and visual pose
	estimation using GPU accelerated stereo reconstruction.
\end{enumerate}
Mode 1) is designed to be used with off-the-shelf hardware, such as DJI products. The computed map is relying solely on GNSS and heading information to align the images into a global map. However, it is also suited to create a global mosaic of featureless or highly dynamic areas. Mode 2) and higher require a camera with high frame rates suited for visual pose estimation. Outputs in 3) include an estimate of the camera motion, depth maps for each view, an incrementally updated 2D map considering surface elevation (orthophoto), as well as a sequentially expanded dense 3D point cloud of the observed scene.

\section{Related Work}
Computing an orthophoto in real-time involves several steps, spanning a variety of topics. Our approach is strongly relying on visual pose estimation and real-time dense monocular reconstruction, which is why sections \ref{subsec_vslam} and \ref{subsec_densemono} are outlining the current state-of-the-art in these fields. Afterwards in \ref{subsec_rtmapping} other, existing real-time mapping frameworks for UAVs are presented.

\subsection{Visual SLAM}
\label{subsec_vslam}
Simultaneous Localization and Mapping (SLAM) is a general term for techniques that obtain both the structure of an unknown environment and the sensor motion in that environment \cite{Taketomi2017}. It was originally developed for autonomous control of robots, but has found a variety of new applications in e.g. augmented reality or self-driving cars since then. Practicable solutions exist for a wide range of sensors today, such as rotary encoders, laser scanner, GNSS receivers and cameras. Especially the latter has undergone active research in the last years due to the simplicity and flexibility of the hardware setup and is typically referred to as 'visual' SLAM.

One notable work in this context was carried out by Mur-Artal et al. \cite{murTRO2015} with their 'ORB SLAM2' framework. At the time of release it set new standards in the field providing not only a visual tracking module, but also global optimization using loop-closing and a relocalization strategy. ORB SLAM2 relies on distinctive point features like corners and edges in the image (indirect approach), however in featureless regions such are difficult to obtain. Other techniques are aiming to evaluate the full image for pose tracking, allowing more robust and accurate results (direct approaches). Direct Sparse Odometry (DSO) \cite{dso} utilizes a photometric optimization formulation to align images by minimizing the difference in pixel intensities. Though, due to the missing abstraction in form of features it is much more challenging to identify already visited locations without permanently saving whole images. Hybrid approaches like Semi-direct Visual Odomoetry (SVO) \cite{Forster2014ICRA} use the direct image alignment on a set of pixel patches, thus being able to apply existing, coordinate-based global optimization techniques, but still offer robustness in featureless areas. 

All of the frameworks share real-time performance and are available open source, but none of them are developed and tested for a downward pointing camera in an aerial mapping scenario.

\begin{figure*}[ht]
	\centering
	\includegraphics[width=13cm,keepaspectratio]{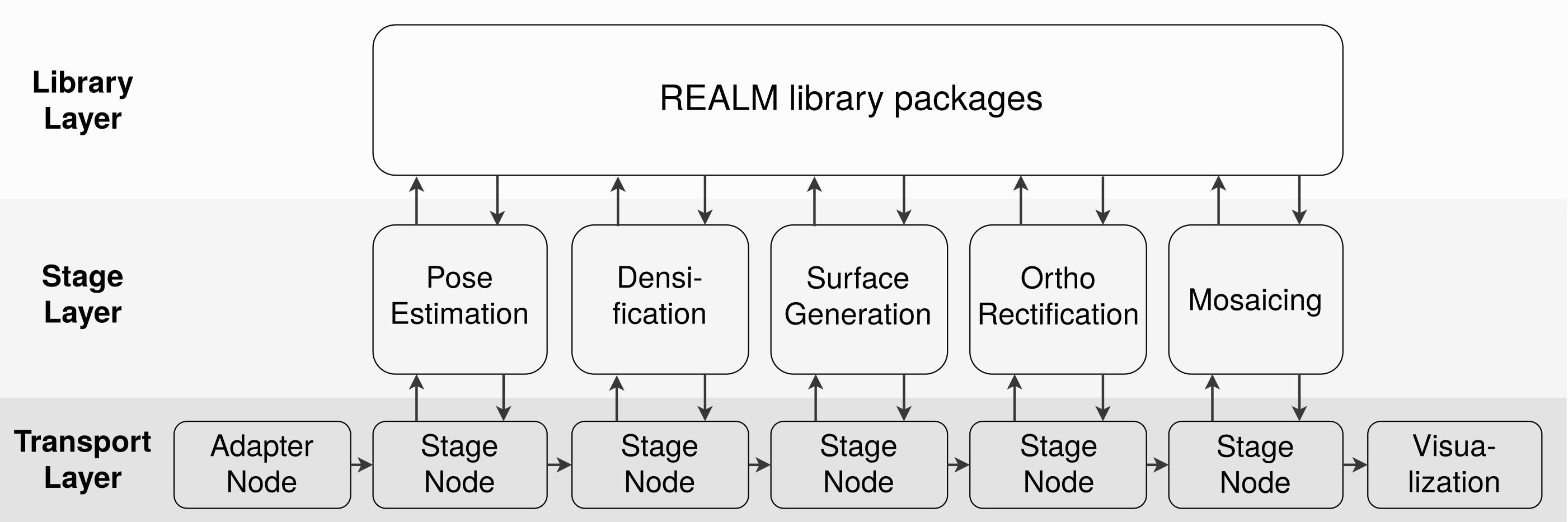}
	\caption{Schematic overview of the proposed implementation. The pipeline consists of three different layers. First is the transport layer, which is realized with ROS. It ensures the data is distributed across all devices in the local, wireless network. Second is the stage layer, which contains the process control and feeds data to the backbone algorithms. The latter are in the library layer. }
	\label{fig:architecture}
\end{figure*}

\subsection{Monocular Dense Reconstruction}
\label{subsec_densemono}

To correct perspective distortions in an aerial map induced by geometric properties of the scene (e.g. tall buildings), information about the surface has to be acquired. Considerable efforts have been made in the past decades to realize monocular dense reconstruction, which is also often known as structure-from-motion in computer vision \cite{Hartley:2003:MVG:861369}, photogrammetry in geodesy \cite{Schelling_VGI_198507} or partially also as monocular SLAM in robotics \cite{Davison:2003:RSL:946247.946734}. As the name suggests, a single moving camera is sufficient to reconstruct dense surface information. While the number of commercial software today is vast (see section \ref{sec_intro}) to perform this reconstruction offline in a multi hour session, a generic product with real-time performance does not yet exist. Companies like Skydio \cite{skydio} just recently pushed the boundaries of what seems possible using deep learning approaches, however as they are closed source not much is known about exactly how they achieve their results and what limitations might exist.

With the real-time requirement the fundamental challenges are very similar to those of visual SLAM. Only that the created map is expected to be much more complete. Featureless regions should not be left out for the sake of a more accurate position. In consequence more data has to be reconstructed and computational load is one of the key challenges for dense surface reconstruction. Considerable open research in this field was published by Pizzoli et al. Their framework 'OpenREMODE' \cite{Pizzoli2014ICRA} provides distance weighting on the camera baseline and a Bayesian formulation on depth uncertainty to minimize noise. At the same time it allows parallelization on a graphics card enabling it to run in real-time. GPU acceleration was also crucial for H\"ane et al. \cite{HaneHLSP14}, who use a fixed number of virtual planes to sweep in different directions across a set of images and identify point correspondences sharing the same plane. Their 'Plane Sweep Library' (PSL) is specifically designed for real-time reconstruction in urban environments due to the naturally 'box-shaped' setting with a limited number of very distinctive planes. However, it will also prove quite performant for aerial mapping.

\subsection{Real-time Mapping}
\label{subsec_rtmapping} 

In comparison to the vast literature about visual SLAM and dense reconstruction, the sources investigating real-time mapping with a UAV are scarce. Traditional approaches perform mainly 2D panorama stitching by detection and matching of feature points between consecutive images, as for example \cite{KEKEC20141755}. However, this strategy mainly relies on computation of the perspective transformation matrix (homography), which in turn only describes the motion between two image planes. This representation lacks flexibility, as it does not allow to use well known techniques from the field of photogrammetry and structure from motion. The limitation to planar surfaces excludes consideration of 3D-data, which in turn would improve the results especially in lower altitudes and with considerable ground elevation. Additionally, almost any use case for real-time aerial mapping would benefit from a digital surface model.

To overcome this fundamental limitation the 3D pose of the acquiring camera is needed. Bu et al. took a step in this direction by publishing their open source software 'Map2DFusion' \cite{DBLP:conf/iros/BuZWL16}. It replaces the image alignment module of a typical stitching pipeline with a state of the art visual SLAM. In consequence, loop closing, global optimization and robust tracking in visual challenging environments is externally provided by a matured, well investigated framework. Starting from the 3D pose images are then projected into a common reference plane to create a global mosaic. Surface elevation, however, is not taken into consideration and visual distortions are inevitable. To achieve what is recognized as aerial photogrammetry today, not only the camera pose must be reconstructed, but also the surface structure. Hinzmann et al. \cite{fsr_hinzmann_2017} were the first to provide a full pipeline from pose estimation, over dense scene reconstruction to orthomosaic generation in real-time. Though, they are neither utilizing state-of-the-art visual SLAM for pose estimation, nor provide an interface to integrate other dense reconstruction implementations. The framework presented in the next few chapters is aiming for both, while its architecture is universal enough to perform simple stitching as well as the full 3D reconstruction.

\section{OpenREALM}
\label{sec_openrealm}

In this chapter an implementation for real-time aerial mapping is proposed. Major contribution of the framework is not to provide specific algorithms or mathematic formulations, but to lay out a fundamental architecture that is versatile and robust to use.

Figure \ref{fig:architecture} shows the overall design in three different layers. The first is the 'Transport Layer' realized with Robot Operating System (ROS). In the aerial mapping scenario there are at least two different processing devices. One device is onboard the UAV and wired to the camera to save or distribute the acquired images. The other is typically the Ground Control Station (GCS) operated by a user to plan, execute and monitor flight missions. In our case, communication between those two devices needs substantial bandwidth to transmit image files, but both can also share processing workload. ROS is ideal for this as it distributes data via WiFi in a modular publisher-subscriber pattern. Captured data consists of an image with Geo-Tag and heading information of the camera. These are brought into the core framework with an adapter node, that might be customized to the specific application. We preferred to simply read and write the required data into and from the image's meta information as Exiv2 tags.

The second layer is the 'Stage Layer' and contains the process control algorithms. Because the proposed architecture is designed as pipeline, every  stage is encapsulated and exchanges results only in one direction through its corresponding ROS node. That way outputs of one stage are consequently inputs of the following. This means some limitations, for example missing feedback loops between the stages, but has the advantage of a clear design and strict task assignment. Individual stages hold the key technology and are described furthermore in detail.

\vspace{0.3cm}
\begin{figure}[h]
	\centering
	\includegraphics[width=\linewidth,keepaspectratio]{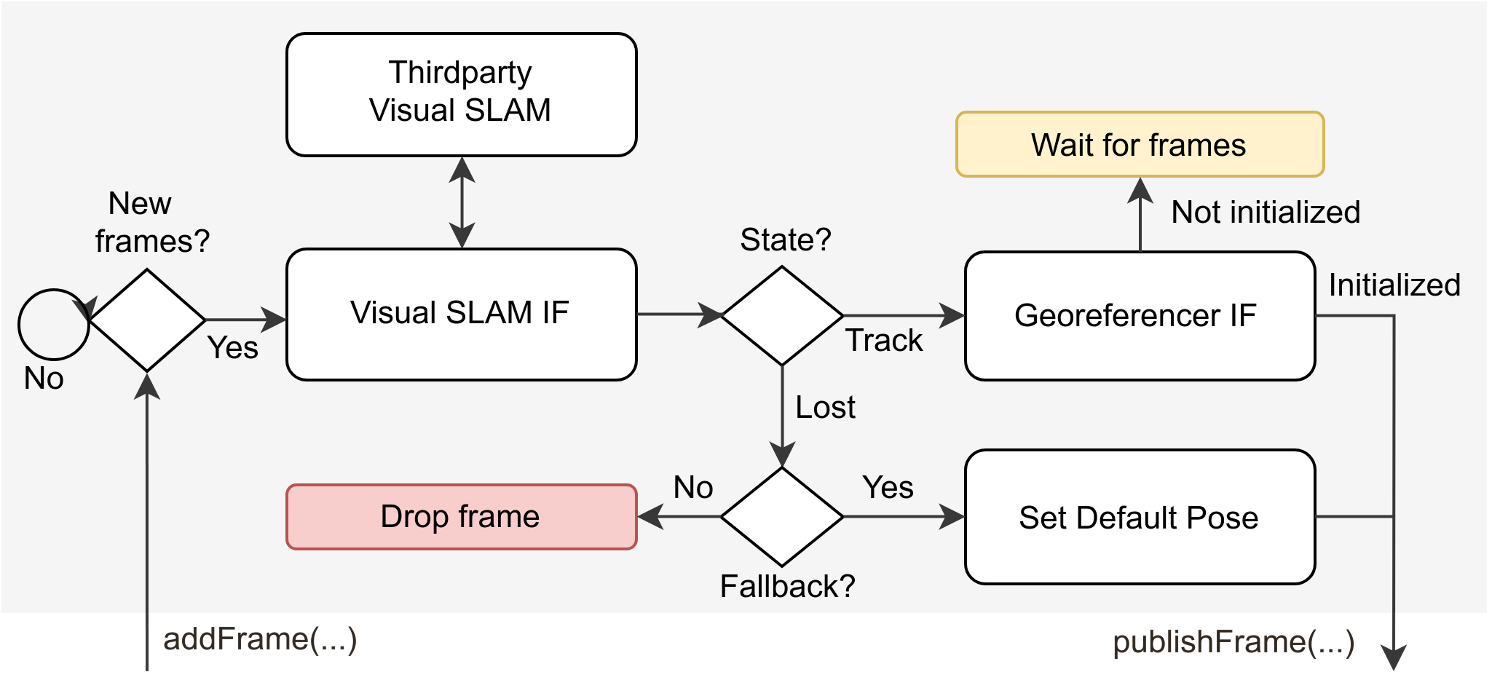}
	\caption{Basic structure of the pose estimation stage. Notable is the visual SLAM interface, which allows integration of state-of-the-art frameworks, and the georeference interface to transform the visual coordinate frame into a global, geographic system.}
	\label{fig:pose_est}
\end{figure}

\subsection{Pose Estimation}
\label{subsec:pose_est}

The schematic workflow of the pose estimation stage is displayed in fig. \ref{fig:pose_est}. On start, it creates a visual SLAM interface (IF) for one of the supported frameworks. This can currently either be ORB SLAM 2, DSO or SVO, depending on the initial arguments passed. Incoming frames are redirected by the interface to the actual SLAM implementation, where the camera pose matrix $M$ is estimated. Note that $M$ is defined as transformation from the camera to the world frame. If tracking was successful, frames are processed in the so called 'georeferencer'. This module tries to identify the transformation from the local, visual to the global, geographic coordinate system utilizing visual and GNSS position. Since an initial set of measurements is necessary for a robust computation, incoming frames might be queued until the estimated error is below a certain threshold. After solving for the arbitrary scale and aligning visual and GNSS trajectory, all frames are published. After this step the computed pose is described as
\begin{equation}
	M =
	\begin{pmatrix}
	r_{11} & r_{12} & r_{13} & t^{C}_{E} \\
	r_{21} & r_{22} & r_{23} & t^{C}_{N} \\
	r_{31} & r_{32} & r_{33} & t^{C}_{Alt} \\
	\end{pmatrix},
\end{equation}
where $r_{ij}$ are the components of the (3x3) rotation matrix $R$, $(t^{C}_{E}, t^{C}_{N})$ UTM coordinates of the global camera position and $t^{C}_{Alt}$ the relative altitude of the camera. 

In case the visual SLAM framework is not able to track the current frame due to e.g. featureless surfaces like water or plane fields, the state switches to 'Lost' and no visual pose is set. To avoid a complete failure of the mapping process in this scenario, a fallback solution can be computed with
\begin{equation}
	M_{Default} =
	\begin{pmatrix}
	cos(-\phi) & -sin(-\phi) & 0 & t^{UAV}_{E} \\
	sin(-\phi) & cos(-\phi) & 0 & t^{UAV}_{N} \\
	0 & 0 & 1 & t^{UAV}_{Alt} \\
	\end{pmatrix},
\end{equation}
where $\phi$ is the magnetic heading of the UAV, $(t^{UAV}_{E}, t^{UAV}_{N})$ UTM coordinates of the UAV's GNSS position, $t^{UAV}_{Alt}$ the relative altitude of the UAV. This substitute pose can only be assumed, if the camera is facing the ground, is stabilized around x- and y-axis and its visual and GNSS position are approximately the same.

\subsection{Densification}
\label{subsec:densification}

\begin{figure}[H]
	\centering
	\includegraphics[width=\linewidth,keepaspectratio]{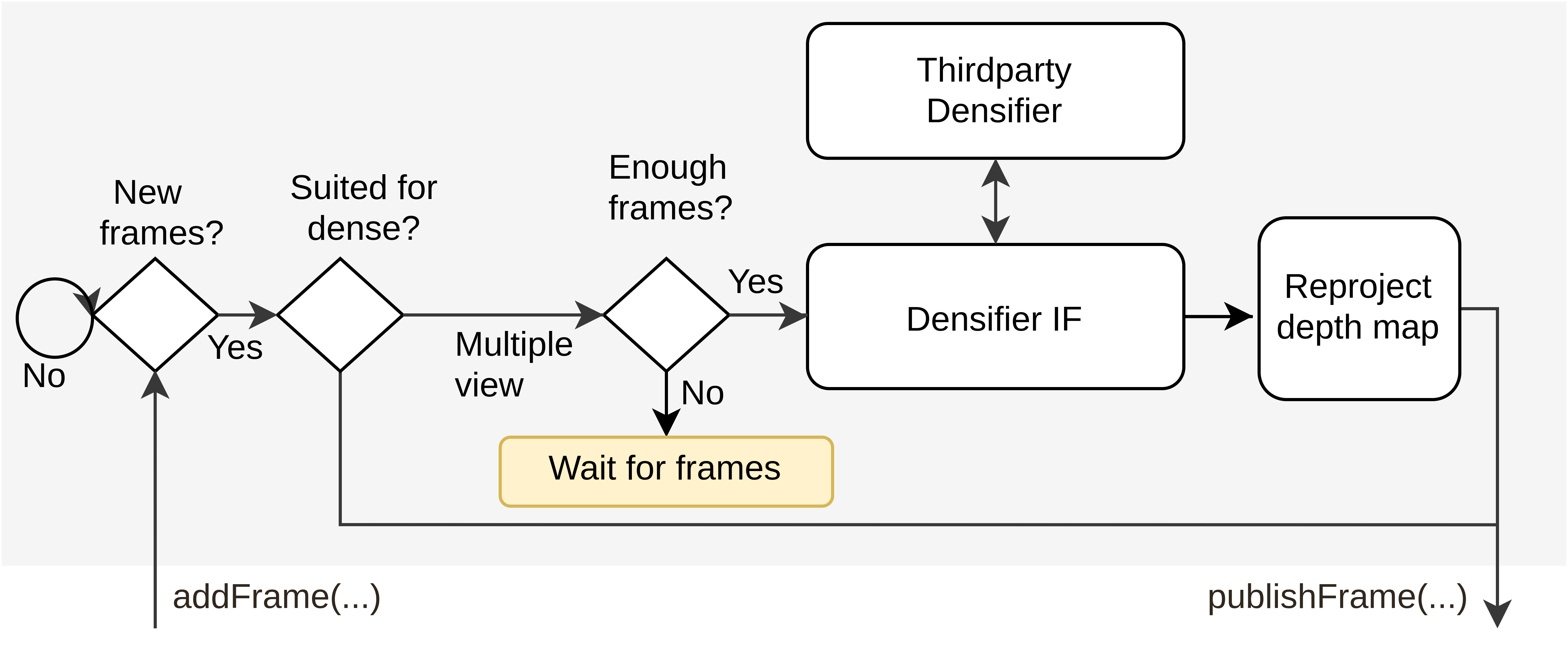}
	\caption{Schematic overview of the densification workflow. Key aspect here is the interface class provided to incorporate existing reconstruction frameworks.}
	\label{fig:dense}
\end{figure}

In the previous stage the camera pose for the current input frame was computed in a geographic coordinate system. This pose can either be visually estimated or based on GNSS and heading information only. The former achieves high accuracy, but is lacking robustness in featureless regions. The latter on the other hand is usually always computable but uncertainties are high and attitude is fixed. In the densification stage only frames with a visually estimated pose are used to reconstruct dense 3D point clouds of the observed surface following the workflow in fig. \ref{fig:dense}.

First, the input frame is checked for suitability. If the pose is identified as visually estimated, depth map creation is initialized. A set of frames depending on the chosen implementation is passed to the densifier interface. This in turn provides, analogous to the visual SLAM interface in the pose estimation stage, the possibility to integrate state of the art reconstruction frameworks. Currently only PSL is incorporated (see \ref{subsec_densemono}). After the dense reconstruction the depth map is projected into a 3D point cloud and any previously existing sparse points are overwritten.

\vspace{0.3cm}

\subsection{Surface Generation}
\label{subsec:surface_gen}

In the previous stages for all processed frames a georeferenced pose was estimated. This could either be based on the image data using visual SLAM, or based on the a-priori GNSS position and heading. Afterwards, only for those frames with visually estimated pose depth maps were generated and projected into dense point clouds. A subsequent stage therefore has to handle the three different types of frames that might occur:
\begin{itemize}
	\item Frames with GNSS position, fixed attitude and no point cloud (if visual pose
	estimation failed),
	\item Frames with visually estimated, accurate pose and sparse point cloud (if only
	densification failed),
	\item Frames with visually estimated, accurate pose and dense point cloud (if all previous
	stages were successful).
\end{itemize}
Surface generation stage evaluates all data of the current input frame and proposes a digital surface model, which describes the observed scene either with a simple plane or a 2.5D elevation map. But it is not only elevation that is of interest, also surface normal and observation angle can usually be reconstructed in the same process. For that purpose an efficient structure is beneficial that tightly couples a specific, geographic location with all information collected about it. Hinzmann et al. proposed the open source library "Grid Map" by P\'{e}ter Fankhauser \cite{Fankhauser2016GridMapLibrary} for such task. It is defined by a region of interest and a ground sampling distance (GSD). Multiple layers of data can be stacked so that every cell of the grid consists of a multidimensional vector of information. For OpenREALM we adopted this idea, but reimplemented some modules. Because Grid Map is designed to be of fixed size and move with a robotic system, dynamically growing the map is rather inefficient. However, in the later stages of our framework exactly this will be useful (see section \ref{subsec:mosaicing}).

Algorithm \ref{alg:planar_surfaces} shows the general workflow in case an incoming frame was identified as 'planar'. The rough ground dimensions must be known and can be computed by projecting the frame into a common reference plane. In the next step a grid is created and filled with a surface of zero elevation. Because the map is later resized, GSD in this special case is not important as long as the structure contains at least one cell.

\begin{algorithm}[t]
	\caption{DSM Creation for Planar Surfaces}
	\label{alg:planar_surfaces}
	\begin{algorithmic}[1] 
		\Require Region of interest for current frame (ROI)
		\State GridMap map('elevation', 'valid')
		\State map.setGeometry(ROI, GSD = 1.0)
		\State map.addLayer('elevation', zeros)
		\State map.addLayer('valid', all)
	\end{algorithmic}
\end{algorithm}

\begin{algorithm}[t]
	\caption{DSM Creation for Elevated Surfaces}
	\label{alg:elevated_surfaces}
	\begin{algorithmic}[1] 
		\Require Region of interest for current frame (ROI), \\ 
		\hspace{0.5cm}dense point cloud
		\State \textbf{KdTree} kdtree = initKdTree(dense cloud)
		\State \textbf{double} resolution = \\
		\hspace{0.5cm} estimateNearestPointDistance(kd tree, dense cloud)
		\State map.setGeometry(ROI, GSD = resolution)
		\State map.addLayer('elevation', zeros)
		\State map.addLayer('valid', none)
		\For{every cell in map}
			\State \textbf{Point} query point = $(x_{cell} , y_{cell})$
			\State \textbf{vector$<$Point$>$} neighbours = 
			\State kdtree $\rightarrow$ findNearestNeighbours(query point)
			\If{neighbours found}
				\State map.at($x_{cell}$ , $y_{cell}$ , 'elevation') = \\ 
				\hspace{1.5cm} interpolateHeight(neighbours)
				\State map.at($x_{cell}$ , $y_{cell}$ , 'valid') = true
			\EndIf
		\EndFor
	\end{algorithmic}
\end{algorithm}

In contrast, the creation of the elevated surface is more complex. Algorithm \ref{alg:elevated_surfaces} shows the implementation as pseudocode. It follows mainly the workflow presented by Timo Hinzmann et al. \cite{fsr_hinzmann_2017}. First the x- and y-coordinates of the dense cloud are structured by a 2-dimensional, binary k-d tree. In the next step this k-d tree is used to compute the nearest neighbour distance for 1$\%$ of all points. The result is assumed as GSD for the subsequent grid map creation, while the region of interest is again provided by the projection of the frame into the reference plane. After adding the layers 'elevation' and 'valid', a nearest neighbour search for every cell $(x_{cell}, y_{cell})$ of the grid is carried out. Note, that $x_{cell}$ and $y_{cell}$ are both UTM coordinates. For all detected neighbours the z-component is extracted from the dense cloud and the resulting height of the cell is finally interpolated.

\subsection{Ortho Rectification}
\label{subsec:ortho_rect}

Goal of the ortho rectification stage is to use the previously estimated surface model and camera pose to rectify the visual distortion of the image induced by the viewing angle and surface structure. At best, the resulting orthophoto is in high resolution so points of interest (e.g. humans, cars, ...) can be easily detected. Hinzmann et al. presented two different approaches to achieve such correction:
\begin{enumerate}
	\item Point Cloud-Based Orthomosaic (Forward Projection)
	\item Grid-Based Orthomosaic (Backward Projection)
\end{enumerate}
While method 1) had the lowest computation time, the authors remarked the strong dependency on the reconstructed dense cloud. Especially the smaller area coverage due to holes in the point cloud were noted. But there are more reasons to consider method 2). By saving the elevation of the observed scene as 2.5D grid map, it can be treated as a regular single channel image with floating point data instead of intensity values. Therefore it can also be efficiently resized to whatever resolution is necessary, just like a regular image. In conclusion the spatial and texture resolution can be treated as two separate parameters. The spatial resolution mainly depends on the densification and surface generation stages, the texture is independent and can be set to a value of choice only limited by the raw image resolution. This is especially useful, if the input images of the mapping pipeline are significantly larger than the multiple view reconstruction can process. For this implementation therefore the 'Grid-based Orthomosaic' technique was chosen.

\begin{figure}[t]
	\centering
	\includegraphics[height=5cm,keepaspectratio]{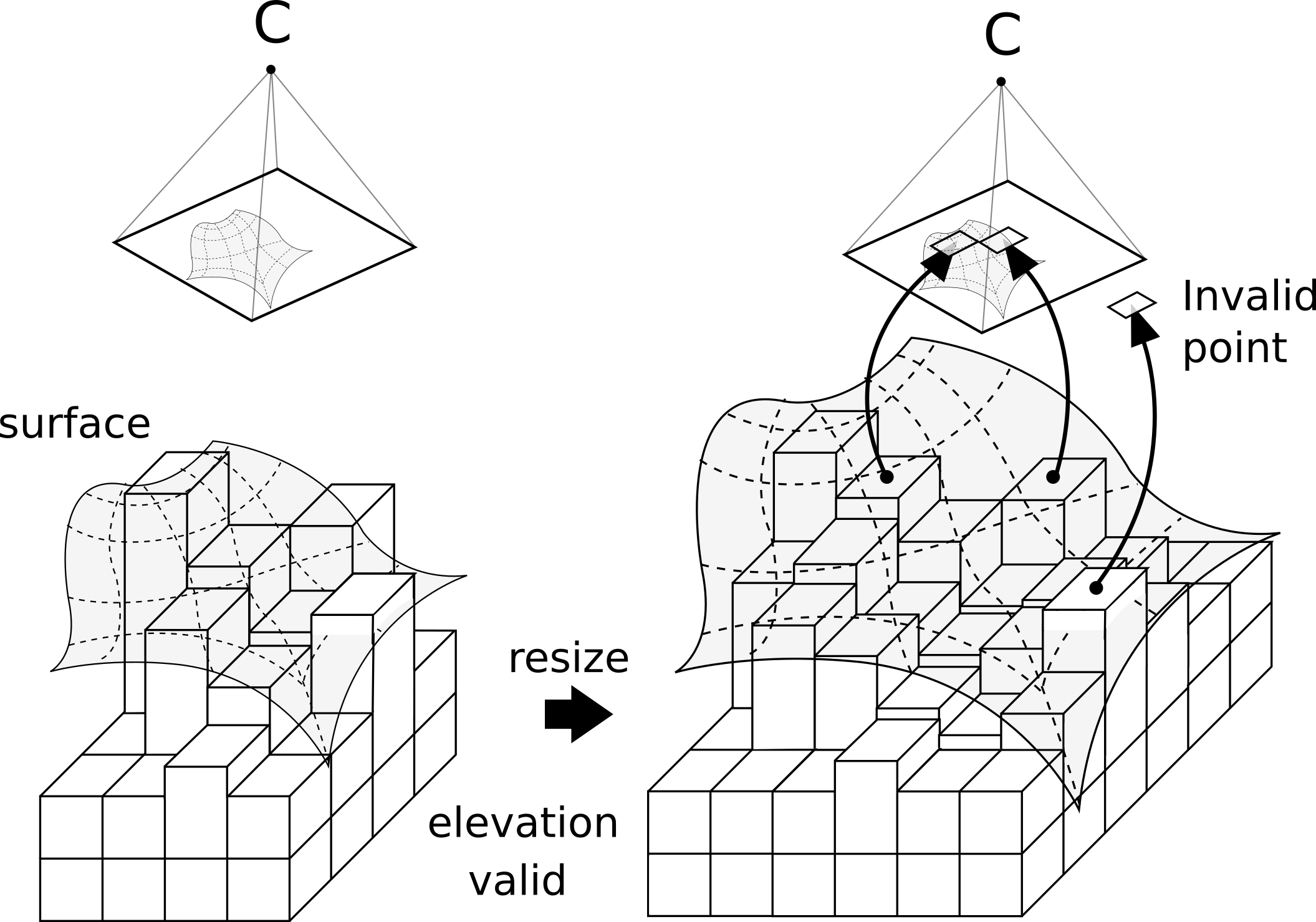}
	\caption{During 'Ortho Rectification' the initial grid map is resized to the maximum desired GSD. That way spatial and texture resolution will become independent parameters, which potentially increases the level of detail in the final orthomosaic.}
	\label{fig:ortho_rect}
\end{figure}

Fig. \ref{fig:ortho_rect} visualizes the basic workflow of the ortho rectification. First, the input grid map containing elevation and validity layer will be resized to the desired orthophoto ground sampling distance. In the next step a 3D point X will be created for each cell of the grid with
\begin{equation}
	X=(x_{cell} , y_{cell} , h_{cell})^{T} ,
\end{equation}
where $(x_{cell}, y_{cell})$ the coordinates and $h_{cell}$ the elevation in the grid at a specific position represent. This step is followed by a back projection into the camera image according to
\begin{equation}
	x = KRX + t,
\end{equation}
with $K$ the camera calibration matrix, $R$ the 3-dimensional rotation, $t$ the translation vector of the camera pose and $x=(u, v, 1)^{T}$ containing the pixel location $(u, v)$ in homogenous coordinates.
Note, that in case of the planar surface assumption the math does not change, $h_{cell}$ will just be zero for all grid cells. 

The observed pixel position $(u, v)$ for the specific cell is now determined and can be set inside a new color layer containing RGB information. Due to noise in the elevation map or a weak pose estimation back projected points may be identified as outside the image boundaries. These points will be consequently marked as invalid. Besides the rectification of the image, the angle of observation is computed for every cell during this stage. It is an additional parameter to achieve high orthogonality in the final mosaic.

\begin{figure}[t]
	\centering
	\includegraphics[height=6cm,keepaspectratio]{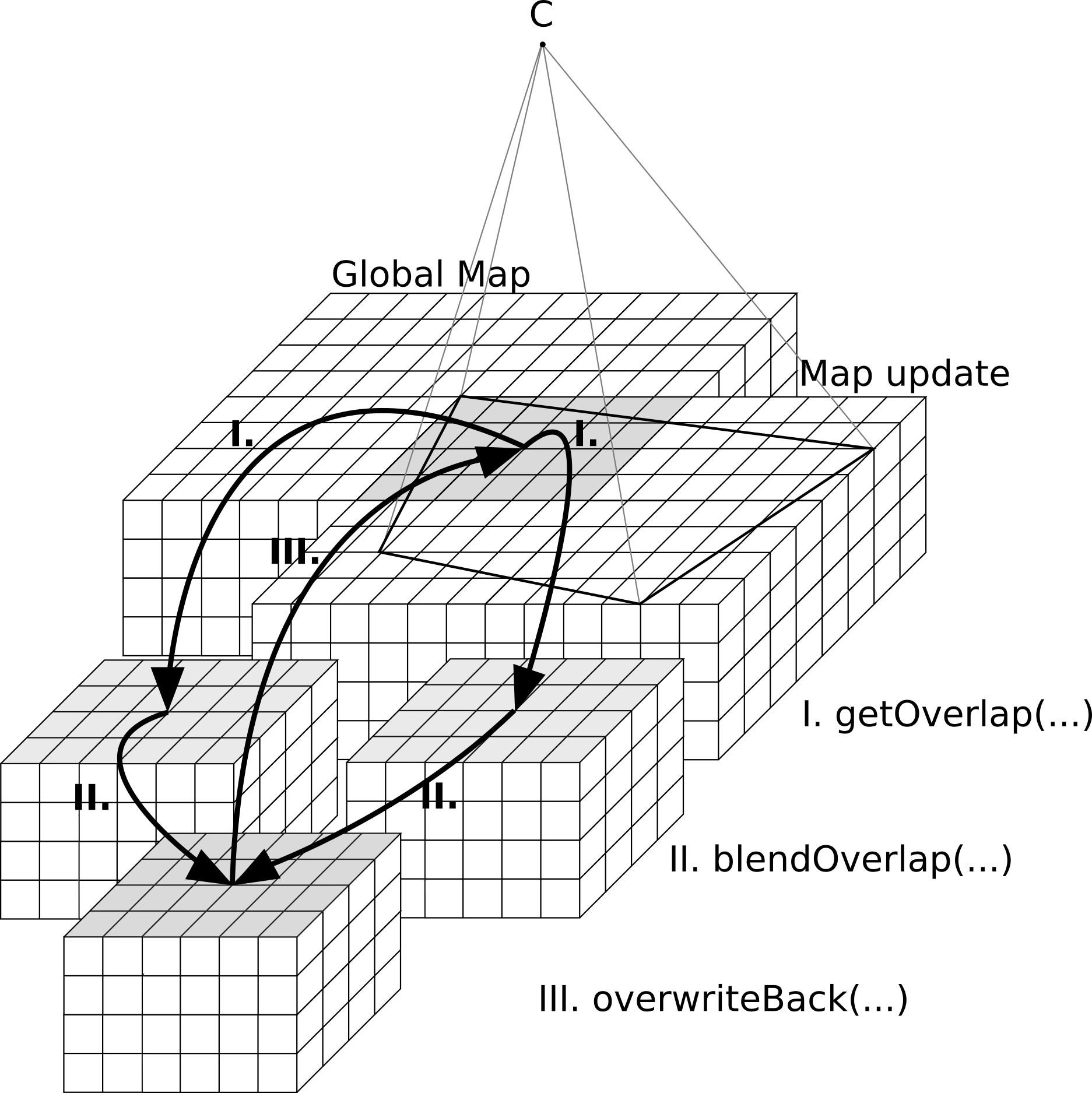}
	\caption{For the mosaicing stage incremental map updates need to be blended together to achieve a consistent global solution. Overlapping regions are characterized by two different sets of information about the same area, which requires an efficient and fast fusion strategy.}
	\label{fig:mosaic}
\end{figure}

\subsection{Mosaicing}
\label{subsec:mosaicing}

Mosaicing is the final processing stage and fuses all previously collected data into a single scene representation. While all prior stages are able to keep the computing resources roughly constant over time, mosaicing does not. All sequentially densified, reconstructed and rectified frames are composed into a high resolution mosaic. The main challenge is therefore to keep the required resources minimal. In fig. \ref{fig:mosaic} the workflow is visualized. By receiving the first frame from the ortho rectification stage the global map is initialized. Afterwards new frames are referred to as 'map update' and can be separated into regions with no prior information or overlapping regions. The former are directly written into the global map. The latter however are extracted so that two submaps are present, both describing only the overlap for the respective data (either global map overlap region or map update overlap region). In the next step one blended value which best describes the surface in each layer is computed for each grid element of the submaps. This blended region is finally written back in the global mosaic.

For the blending of the grid cells a variety of strategies can be applied. In this paper a probabilistic approach was chosen and is described in the following. The underlying problem statement can be summarized as: 'If two different hypothesis for the elevation of a grid cell exist, which one is chosen?'. To decide this basic question, three additional layers are added to the global map, the 'elevation variance', 'elevation hypothesis' and the 'number of observations'. As soon as a map update arrives, for every cell of the overlapping region a temporary, floating average of the new elevation is computed with
\begin{equation}
	\hat{x}_{ij}=\frac{n_{ij}}{(n_{ij}+1)}\hat{x}^{global}_{ij}+\frac{1}{(n_{ij}+1)}x^{update}_{ij}
\end{equation}
where $\hat{x}^{global}_{ij}$ is the current averaged elevation in the global map, $n_{ij}$ the number of observations of $\hat{x}^{global}_{ij}$ and $x^{update}_{ij}$ the elevation of the map update. In a similar way the floating sample variance $s^{2}_{ij}$ is estimated with
\begin{equation}
s^{2}_{ij}=\frac{(n_{ij}-1)}{n_{ij}}{s^{global}_{ij}}^{2}+\frac{(x^{update}_{ij}-\hat{x}^{global}_{ij})^{2}}{(n_{ij}+1)}.
\end{equation}
If $s^{2}_{ij}$ is below a certain threshold, the new values for variance and average are written to the grid map layers. Additionally, the number of observations for the current cell is incremented. In case $s^{2}_{ij}$ exceeds the variance threshold, two different hypothesis exist for the specific cell. After the first appearance of such a new hypothesis, it can not be resolved and is therefore written into the 'elevation hypothesis' layer, while the existing data is kept as is. As soon as a new update for the specific cell arrives in the mosaicing stage, the floating average and sample variance are compared to both, the elevation set and the possible second hypothesis. Now the one with the lower sample variance is selected as the most likely one, while the other is written to the hypothesis layer. The above strategy aims to reduce noise in the elevation by minimizing its variance.

\section{Evaluation}
\label{sec:evaluation}

A comprehensive analysis of the proposed framework is out of the scope of this paper. The focus is therefore set on evaluating if the initial requirements are met. Key feature is to provide a 2D map in real-time. Because creating an orthophoto based on a digital surface model is the most challenging task, this is considered the default mode of operation in the following evaluation. However, for the sake of comparison in section \ref{subsec:quality_orthophoto} we also show the mapping results for the other modes.

\subsection{Dataset}
\label{subsec:dataset}

To the best of our knowledge there are no existing, public datasets for benchmarking aerial mapping frameworks, that suite visual SLAM applications (high framerate, high quality camera with fixed intrinsics). Consequently, it was necessary to create one. Our custom dataset was acquired by a 560mm wheelbase quadrotor with 2 kg take-off weight and an estimated total flight time of 15 minutes. The UAV navigation stack consists of a Pixhawk 2.1 autopilot running APM with a ublox NEO-7 GNSS module and HMC5883L digital compass. It is equipped with an Odroid XU4 companion computer that is wired to the Pixhawk. The Odroid runs Mavros and a custom camera node to grab images, create Exiv2 tags (e.g. GNSS and heading information) and write all data to the harddrive. Both, image stream and global position, are updated with 10 Hz but are not synchronized. The vision setup consists of a UI-5280CP Rev. 2 manufactured by IDS Imaging Development Systems GmbH with global shutter, 5 MPix resolution and an image size of 2456x2054 pixels. However, for the sake of bandwidth reduction and increased performance of the companion computer the images were subsampled to 1228x1027 pixels. As processing hardware for evaluation all tests were performed on a XMG P406 laptop with Intel i7-6700HQ cpu, NVIDIA GeForce GTX 970M graphics card and 16 GB RAM.
\begin{figure}[t]
	\centering
	\includegraphics[width=\linewidth,keepaspectratio]{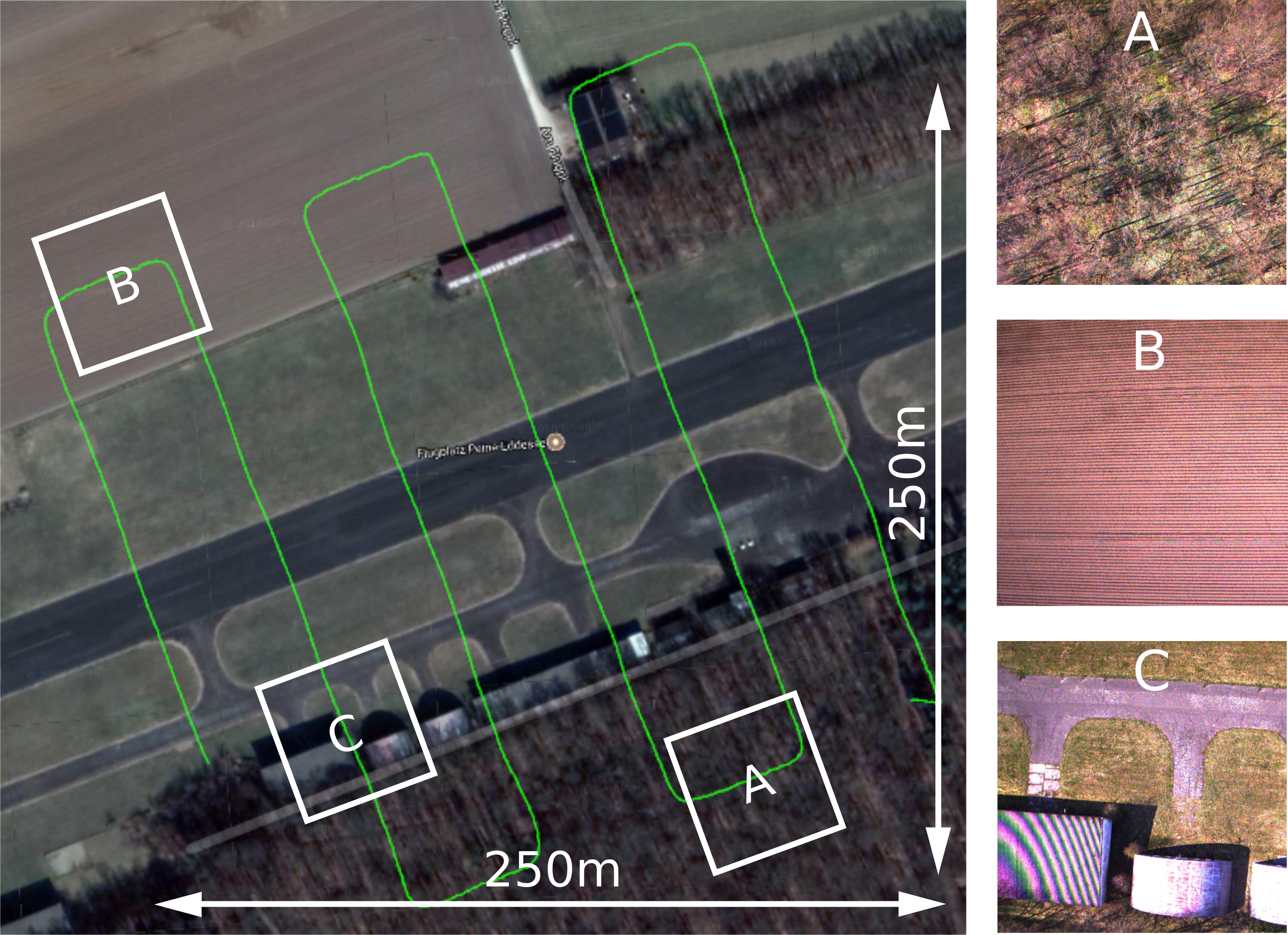}
	\caption{Overview of the acquired custom dataset. In A,B and C details are shown, that are challenging for visual algorithms due to their lack or ambiguity of features.}
	\label{fig:scene}
\end{figure}
With the previously stated hardware a set of 3276 images was acquired. The UAV trajectory described serpentines with 99$\%$ front and 50$\%$ side overlap above the observed scene (fig. \ref{fig:scene}, green), which has the size of roughly 250x250m and shows an abandoned airport in Edemissen, Germany. The dataset has some challenges for visual algorithms, which are also displayed. In A) the forest in the south can be seen, that has very repetitive texture which hardly allows to extract unique points for feature-based visual SLAM. Same goes for B), but induced by the homogenous surface. Illustration C) in turn reveals a regional aliasing effect on the corrugated iron caused by the fact that the repeating roof pattern is in the realm of the GSD.

\subsection{Pose Accuracy of the visual SLAM}
\label{subsec:pose_est_perf}

\begin{figure}[h]
	\centering
	\begin{subfigure}[b]{\linewidth}
		\centering
		\includegraphics[width=7.5cm,keepaspectratio]{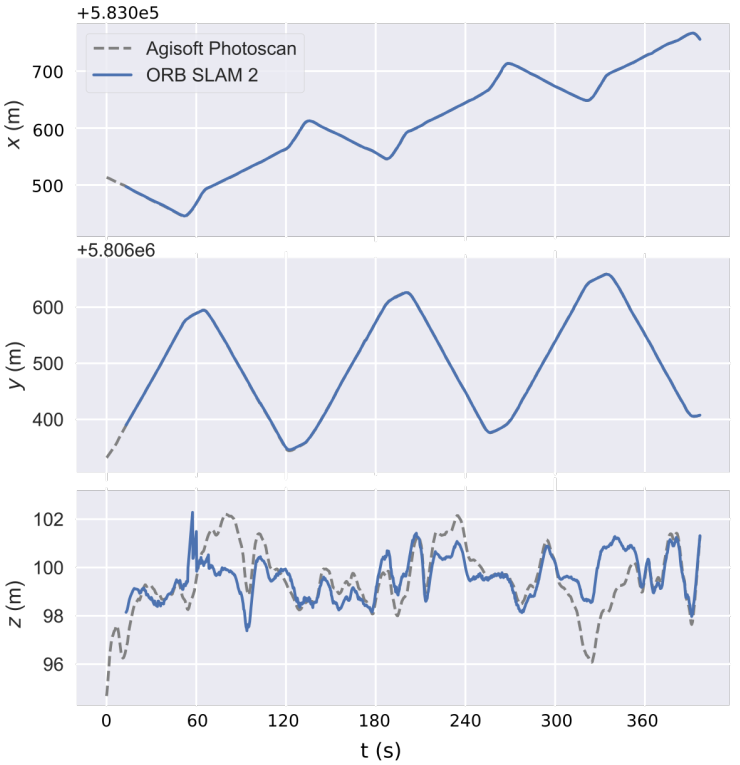}
		\caption{Trajectories plotted for ORB SLAM2 (blue) and Agisoft Metashape (dotted). The overall alignment is good, especially in the x- and y-axis. The highest difference can be found in the z-axis, which represents the reconstructed depth.}
	\end{subfigure}
	\begin{subfigure}[b]{\linewidth}
		\centering
		\includegraphics[width=7.5cm,keepaspectratio]{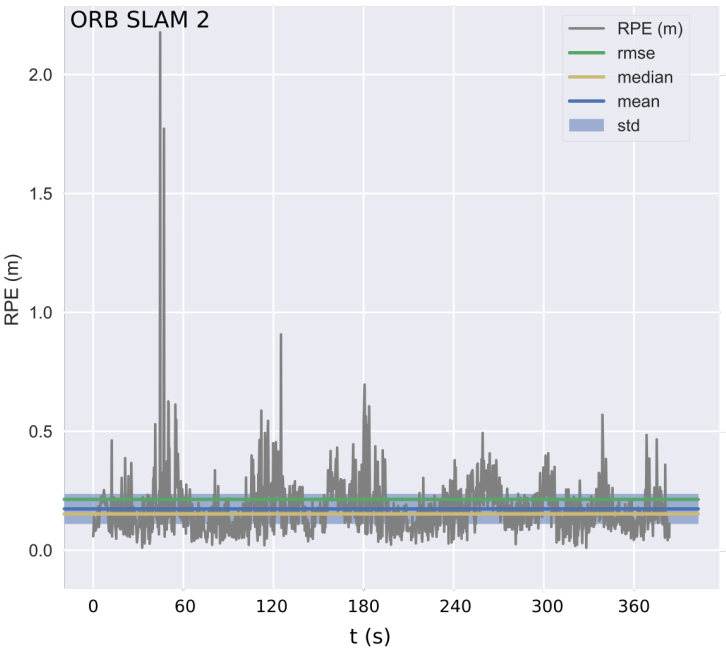}
		\hfill
		\caption{Relative and absolute pose error (RPE/APE) for the ORB SLAM2 approach over the full dataset.}
	\end{subfigure}
	\caption{Comparison of pose accuracy for ORB SLAM2. The ground truth trajectory was computed using Agisoft Metashape.}
	\label{fig:perf_orbslam}
	\vspace{-0.5cm}
\end{figure}

At first an evaluation of the pose estimation should be carried out. A high pose accuracy is important, because all consecutive processing stages will inherit the uncertainties. As ground truth we computed the trajectory for all 3276 images with the classical, offline photogrammetry software Agisoft Metashape. As visual SLAM framework we limit our analysis to ORB SLAM2. In fig. \ref{fig:perf_orbslam} the results are shown. In a) the translational error for every axis is displayed. Especially the xy-alignment has almost no visible displacement. The z-axis on the other hand is equatable to the estimated depth of the scene, which is reconstructed using multiple view geometry. This axis should therefore be the most uncertain one, and indeed a slight deviation to the ground truth can be measured. Between 60-100s the first turning point of the UAV occurs, which is also the visual obstacle B in fig. \ref{fig:scene}. The low number of features in this region reduces the overall accuracy measurably. The relative pose error (RPE) in fig. \ref{fig:perf_orbslam} b) supports this thesis, as it has its highest peak before the turn, and quickly decreases afterwards. An absolute pose error (APE) of 0.53m averaged shows the overall good performance of the ORB SLAM 2 for visual pose estimation.

\subsection{Quality of the Surface Reconstruction}
\label{subsec:quality_surface_rec}

\begin{figure*}[t]
	\centering
	\includegraphics[width=14cm,keepaspectratio]{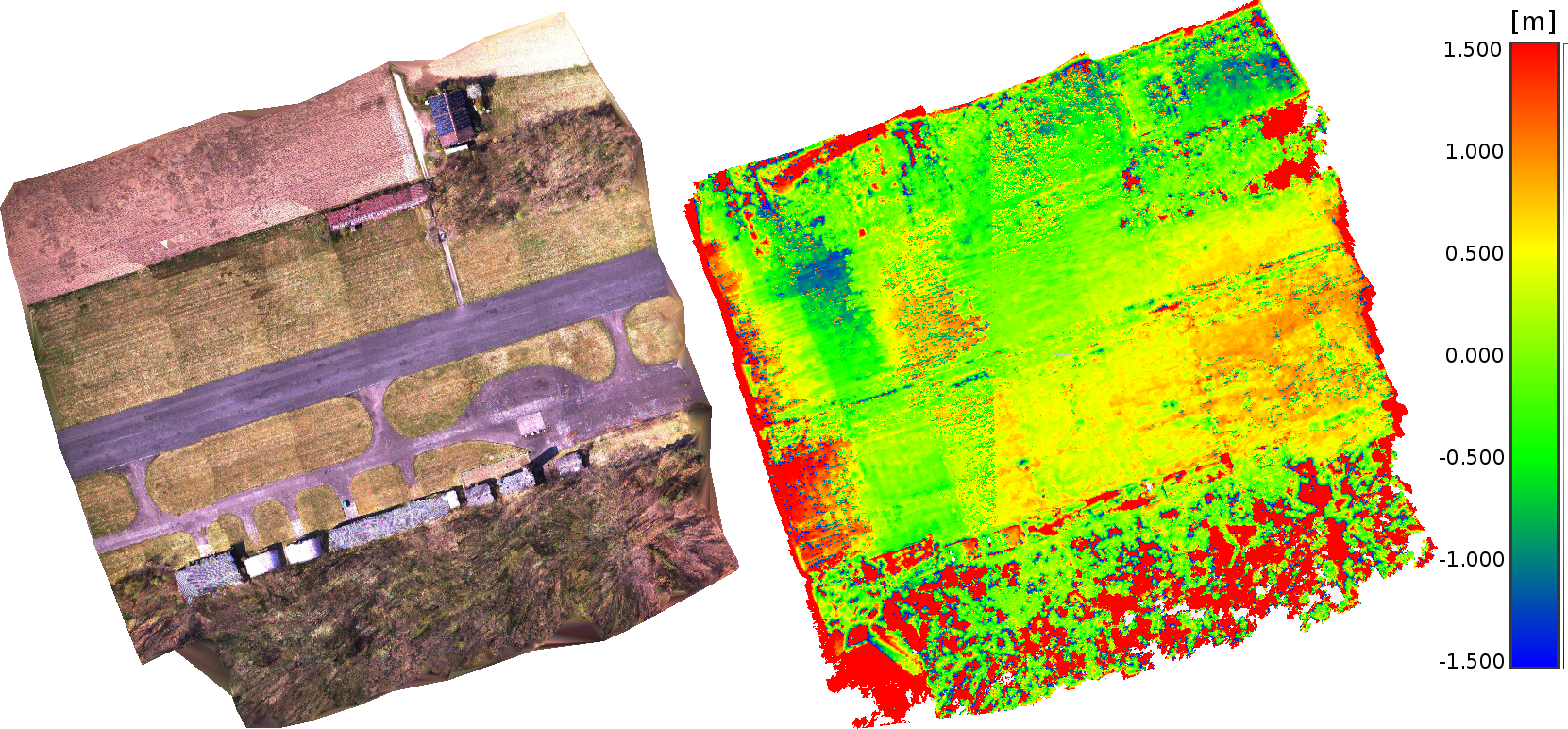}
	\caption{The reconstructed mesh by OpenREALM is displayed on the left. For the sake of evaluation ground truth was computed with Agisoft Metashape and the resulting surface deviations color coded. Red areas in the borders regions are a consequence of missing data. }
	\label{fig:mesh_comp}
	\vspace{-0.2cm}
\end{figure*}

The previous section showed how the visual pose estimation is performing. Considering the challenging, low feature textures in some regions the results are promising. In the next step the 3D surface reconstruction is analyzed. Agisoft Metashape is used to generate ground truth. For that purpose the front overlap of the dataset is reduced to 80$\%$, so computation time remains reasonable (1:45h). Settings for the stages were chosen, so that processing is still live on the hardware, but the spatial resolution is as high as possible. The resulting dense cloud of the final map was exported to the open source software 'CloudCompare' \cite{cloudcompare}. There, it was aligned to the ground truth using iterative closest point algorithm (ICP) to reduce the influence of the georeference. Afterwards every point of the real-time created dense cloud is projected onto the reference mesh. The resulting distance is encoded by color value and displayed at its respective position on the mesh. This is displayed in fig. \ref{fig:mesh_comp}.

Densification of the proposed implementation was processed using PSL. The achieved GSD is 0.15 m/cell. The aliasing effect in the dataset seen in fig. \ref{fig:scene} c) seemed to have a severe influence on the reconstruction process. The house can barely be identified in the 3D. This is not surprising, as PSL tries to compute a depth value for every single pixel, even though the triangulation is very uncertain. The rest of the runway aligns well with the ground truth. The plane field at the top left of the map, just before the turning point, has the highest deviations (-1.5 to 0.5m). This was also the region where the pose estimation showed a significant offset. It is therefore very likely that an initial error in the first stage has propagated to the mosaicing. All in all the results are improvable, however it is important to keep in mind that processing time for OpenREALM was live, Agisoft Metashape in contrast took several hours.

\subsection{Quality of the Orthophoto}
\label{subsec:quality_orthophoto}

Simultaneously with the surface reconstruction in the previous section a 2D orthophoto was generated. Orthophotos provide a fast overview of the mapped area and became essential to aerial photogrammetry. They are aligned to a global coordinate frame of choice and allow to measure distances in real world scale. Quality evaluation of such is difficult, as visual distortions can easily be spotted by humans, but are more challenging to be detected automatically. For that reason a manual inspection should be sufficient. 

\begin{figure*}[t]
	\centering
	\includegraphics[width=\linewidth,keepaspectratio]{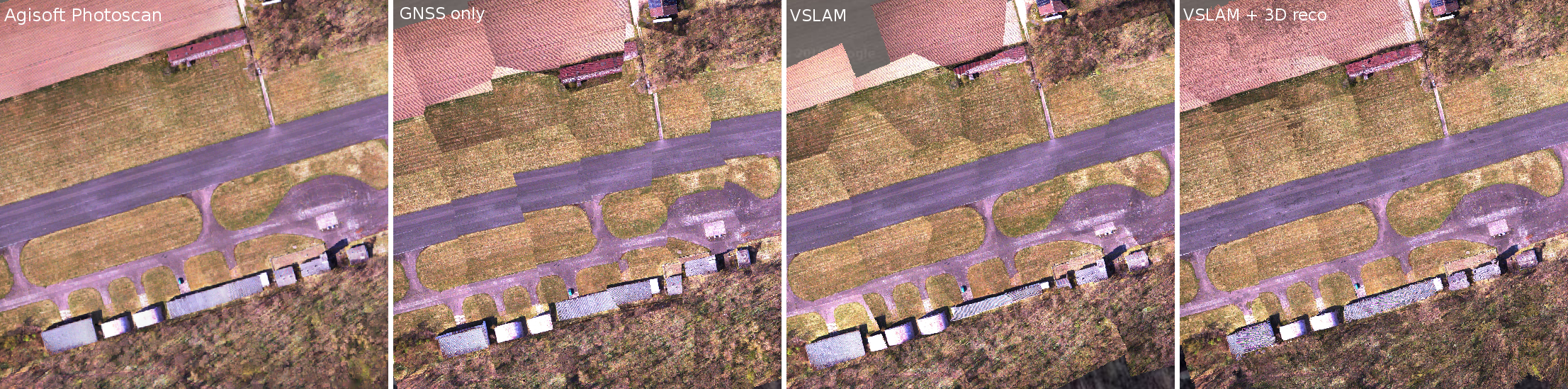}
	\caption{Resulting maps for OpenREALM's different modes of operation compared to the ground truth (left) provided by Agisoft Metashape. Second from the left shows the map generated by aligning the acquired images purely based on the GNSS position and heading of the UAV. On th second from the recond in contrast visual pose estimation using ORB SLAM2 was performed. However, no 3D reconstruction took place. The right figure shows the approach, which provides a 2.5D elevation map as well as true orthophotos.}
	\label{fig:quality_ortho}
	\vspace{-0.4cm}
\end{figure*}

A good orthophoto has high resolution, homogenous appearance, low geometric distortion and no visual artifacts. It must represent the observed scene with as much details as possible and should be accurately aligned globally. Ground truth for the visual inspection is shown in fig. \ref{fig:quality_ortho} (left). As expected, Agisoft Metashape was able to reconstruct the scene without major flaws. The map contains no ghosting or overlapping edges and has a resolution of up to 0.07m/px. Exposure time changes in the camera propagated to the field in the top of the map, which brightened slightly. But it does not influence the integrity of the map significantly.

In comparison, results for all modes of operation are also presented. Second from the left images were positioned solely based on GNSS and heading information. Several obvious misalignments can be seen, that result from a systematic offset between assumed and real camera heading. With synchronization and a stricter alignment of camera and UAV heading the orthophoto for this mode can further be improved. Second to the right visual SLAM was used for pose estimation, but no 3D reconstruction was performed. It represents the state of the art for real-time mapping, as it does not need a full reconstruction of the surface, but allows to create visually appealing results. With improved blending the orthophoto comes close to the ground truth. Though, geometric distortion for both of the modes are a fundamental problem which can not be overcome without surface information. On the right the map with elevation is displayed. It is computationally more expensive due to the densification process, but shows rectified, homogenous results similar to the Metashape orthophoto. Especially the building in the upper part of the map has less artifacts and looks more like the ground truth.

Altogether the surface data improved the global map visually. Larger misalignments and geometric distortions are removed. Though, the greatest benefit is the 3D impression of the observed scene. In use cases like search and rescue such information are crucial for coordination and situational awareness.

\subsection{Real-time Performance}
\label{subsec:rt_perf}

As last step of this section the evaluation of the processing performance is carried out. In general, a stage should process data slightly faster than it receives new input. Due to the fact, that the proposed implementation is designed as multi-threaded pipeline a simple time measurement for each stage however is not sufficient. If for example the densification stage has an average computation time of 0.2s, it might publish new frames for the next stage with 5 Hz, but it might as well publish at 1 Hz. A consecutive stage with an average computation time of 1.0s might then be okay, or overloaded by a factor of 5. By measuring the downtime also idle states are detected. Yet, a downtime of 0.0s might as well mean the stage is running fine. Therefore another approach was chosen that is outlined in the following. The current transport layer is implemented in ROS. Therefore messages exchanged between the processing stages are transferred in the ROS infrastructure. To identify if each stage is performing within the limits the exchanged message rates are tracked. By measuring the input and output message frequency of each stage the overall workload can be estimated. One assumption that is required is, that every stage processes every frame and has no artificial throttle to reduce the output rate. Because of the keyframe selection this is not the case for the pose estimation stage, which is why it is neglected for now. All other stages process and publish every frame they receive. The performance measure is then defined as
\begin{equation}
	\delta_{Perf}=\frac{f_{in}}{f_{out}},
\end{equation}
where $f_{in}$ represents the input and $f_{out}$ the output message frequency. Consequently a stage that is publishing messages as fast as it receives new data $(\delta_{Perf}=1.0)$ is declared as 'real-time performant'. Stages with $\delta_{Perf}>1.0$ are getting frames faster than they are able to publish them, which indicates high workload and is therefore labeled as overloaded. A $\delta_{Perf}<1.0$ should not be possible as such a stage would generate more data than it receives. However, due to the fact that the performance tracker averages the frequency over time it might underlie temporary variations. Therefore only the steady-state is of interest.

\begin{figure}[h]
	\centering
	\includegraphics[width=\linewidth,keepaspectratio]{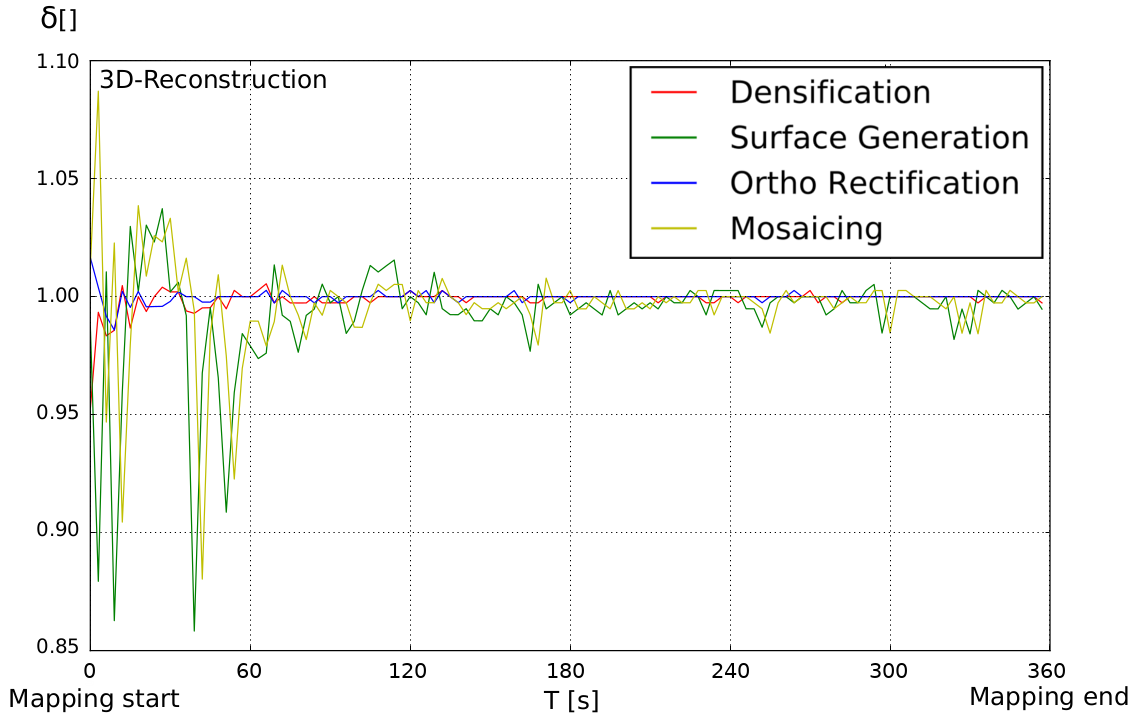}
	\caption{The performance of the individual stages is qualified by the frequency of incoming and outgoing messages. A coefficient of $\delta_{Perf}=1.0$ means, that messages are published as fast as new ones arrive. $\delta_{Perf}>1.0$ indicates an overload, as the stage is not able to keep up with processing the current frame.}	
	\label{fig:rt_perf}
	\vspace{-0.4cm}
\end{figure}

Fig. \ref{fig:rt_perf} shows the performance of the 3D reconstruction approach using PSL. The pose estimation stage is provided with an input image stream of 10 frames per second and outputs keyframes at rougly 2.4 Hz. The resulting performance logs for the other stages expose variations especially in the beginning. The densification stage has recurring peaks. This might be explained by the additional involvement of the GPU or a more complex reconstruction of the scene. Yet, all converge to $\delta_{Perf}=1.0$ and are therefore performing within the requirements.

\section{Conclusion}
\label{sec:conclusion}

We presented a real-time mapping framework for unmanned aerial vehicles. Different modes of operation enable the user to perform GNSS or visual SLAM based image stitching, or to fully reconstruct the 3D surface and extract a geometrically corrected orthophoto. The quality of the surface should further be improved in the future. Current implementations for SLAM and 3D reconstruction are not designed and optimized for an aerial mapping scenario. Further research in this field should therefore improve the results substantially. Also providing a public dataset for benchmarking mapping frameworks might push progress in the future.

%\section*{Appendix}

\vspace{-0.2cm}
\section*{Acknowledgment}

The AeroInspekt project is supported by the BMVI (German Federal Ministry of. Transport and Digital Infrastructure) through the IHATEC Funding Call, Funding code: 19H18006B.

\vspace{0.2cm}
\centering
\includegraphics[width=3cm,keepaspectratio]{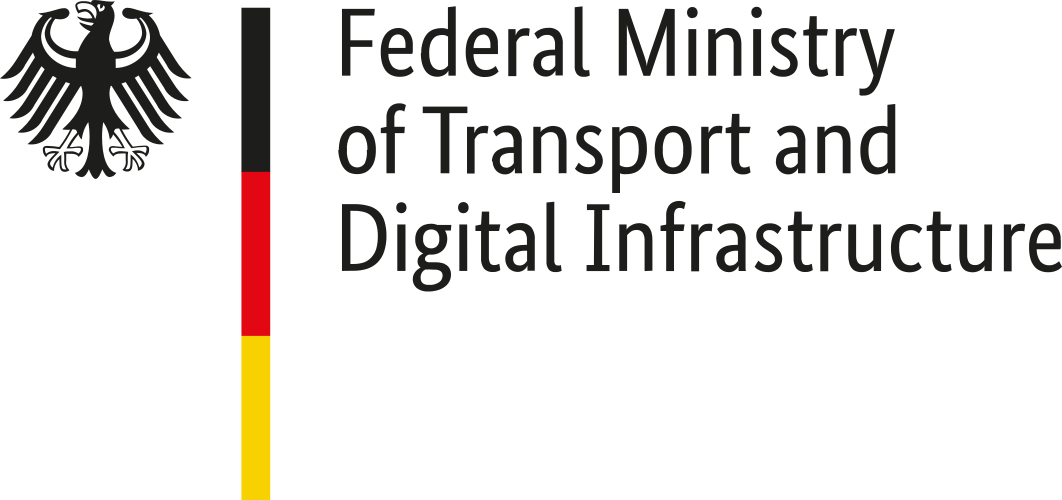}

\renewcommand{\bibname}{References}
\printbibliography

\end{document}